\documentclass[runningheads]{llncs}
\usepackage{graphicx}

\usepackage{amsmath,amsfonts}
\usepackage{algorithmic}
\usepackage{algorithm}
\usepackage{array}
\usepackage[caption=false,font=normalsize,labelfont=sf,textfont=sf]{subfig}
\usepackage{textcomp}
\usepackage{stfloats}
\usepackage{url}
\usepackage{verbatim}
\usepackage{graphicx}
\usepackage{cite}

\usepackage{wrapfig}

\usepackage{amssymb}
\usepackage{booktabs}
\usepackage{xcolor}
\usepackage{bm}
\usepackage{pifont}
\usepackage{enumitem}
\usepackage{soul}
\usepackage{makecell}
\usepackage{tabularx}
\usepackage{gensymb}
\usepackage{siunitx}
\usepackage{balance}

\usepackage{soul}
\usepackage{multicol}
\usepackage{multirow}

\usepackage{etoolbox}

\usepackage[pagebackref,breaklinks,colorlinks]{hyperref}

\def \ie {\emph{i.e.},}
\def \eg {\emph{e.g.},}
\def \etal {\emph{et al.}}

\newcolumntype{Y}{>{\centering\arraybackslash}X}

\newcommand{\tit}[1]{\smallbreak\noindent\textbf{#1.}}

\usepackage[capitalize]{cleveref}
\crefname{section}{Sec.}{Secs.}
\Crefname{section}{Section}{Sections}
\Crefname{table}{Table}{Tables}
\crefname{table}{Tab.}{Tabs.}

\newcommand{\netname}{FourBi}

\begin{document}

\title{Binarizing Documents \\ by Leveraging both Space and Frequency}

\author{Fabio Quattrini\orcidID{0009-0004-3244-6186} \and Vittorio Pippi\orcidID{0009-0001-7365-6348} \and \\ Silvia Cascianelli\orcidID{0000-0001-7885-6050} \and Rita Cucchiara\orcidID{0000-0002-2239-283X}}

\institute{University of Modena and Reggio Emilia, Modena, Italy \\
\email{\{name.surname\}@unimore.it\vspace{-.3cm}}
}

\maketitle              
\begin{abstract}
Document Image Binarization is a well-known problem in Document Analysis and Computer Vision, although it is far from being solved. One of the main challenges of this task is that documents generally exhibit degradations and acquisition artifacts that can greatly vary throughout the page. Nonetheless, even when dealing with a local patch of the document, taking into account the overall appearance of a wide portion of the page can ease the prediction by enriching it with semantic information on the ink and background conditions. In this respect, approaches able to model both local and global information have been proven suitable for this task. In particular, recent applications of Vision Transformer (ViT)-based models, able to model short and long-range dependencies via the attention mechanism, have demonstrated their superiority over standard Convolution-based models, which instead struggle to model global dependencies. In this work, we propose an alternative solution based on the recently introduced Fast Fourier Convolutions, which overcomes the limitation of standard convolutions in modeling global information while requiring fewer parameters than ViTs. We validate the effectiveness of our approach via extensive experimental analysis considering different types of degradations. 

\end{abstract}

\section{Introduction}
Documents with good visual quality are easier to handle when performing Document Analysis tasks, which is why the problems and approaches to Document Enhancement have long been explored. However, this remains a challenging task, especially for old and highly damaged documents, for which automatically achieving good visual quality can benefit efficient content understanding.
Due to preservation conditions and time, these documents exhibit various types of degradations, such as bleed-through ink, ink oxidation, faded ink, smears, scratches, stains, and creases in the paper. Moreover, the digitalization process can introduce artifacts due to image compression and unideal acquisition conditions such as uneven illumination, contrast variation, and blur~\cite{tensmeyer2020historical}. 
A typical Document Enhancement task is Document Binarization, which consists in determining whether each pixel in the document image pertains to ink or to the background. This is a critical step both for facilitating downstream Document Analysis tasks~\cite{stauffer2018keyword, ray2019end,cojocaru2021watch,de2022few,jemni2022enhance, souibgui2022docentr, yang2023gdb,koloda2023context,pippi2023choose,pippi2023evaluating} and for enhancing the readability of the document by the end-users of digital archives and libraries. 
One of the main challenges in Document Binarization is the variability with which the aforementioned degradations appear within the page. For this reason, early-proposed image processing approaches based on estimating and applying a global threshold~\cite{otsu1979threshold,barron2020generalization} have been surpassed by methods working on local document patches, on which an adaptive threshold is calculated and applied. Learning-based solutions, generally based on Convolutional Neural Networks, enhanced the performance by refining the prediction on the single pixel based on its neighbors. 
Nonetheless, a potential drawback of working on document patches is that, in case they contain too different degradations, some artifacts can appear at their borders when restitching them together to form the whole document image. 
For this reason, considering both global and local information has been proven beneficial by the promising performance of hybrid thresholding methods~\cite{wolf2002text}, Image-to-Image translation approaches based on Generative Adversarial Networks (GANs)~\cite{zhao2019document,souibgui2020gan,tamrin2021two}, and multi-scale Convolutional approaches for Segmentation~\cite{tensmeyer2017document,vo2018binarization}. 
Note that global dependencies are difficult to handle with standard Convolution-based models due to the fixed size and locality of their receptive field. 
A strategy to integrate global information is to exploit attention mechanisms over the pixels. This strategy has been successfully applied also for Document Binarization~\cite{souibgui2022docentr,souibgui2022text,yang2023novel}. However, this strategy requires a large amount of training data, which is difficult to collect for document images, 
and usually leads to large models, thus being inefficient. 

To overcome these limitations, we devise an efficient, fully-convolutional model able to integrate global information for Document Binarization by exploiting the Fast Fourier Convolution (FFC)~\cite{chi2020fast}. 
This operator combines spatial and spectral information to expand the receptive field in the frequency domain. 
The benefits of FFCs have been exploited for Inpainting~\cite{suvorov2022resolution}, Detection~\cite{quattrini2023volumetric}, and Super-Resolution~\cite{sinha2022nl} but, to the best of our knowledge, FFCs have not been employed for Document Binarization. However, we argue that their capability to encode both local and global information and model pseudo-periodic patterns and textures is particularly suitable for documents. In fact, text lines have a specific pseudo-periodic pattern, and paper has a very peculiar texture, different from that of artifacts and degradations. 
Moreover, due to their image-wide receptive field in the spectral domain, FFCs are more robust to differences in resolution and scale between training and test images compared to standard convolutions~\cite{suvorov2022resolution}. 
Thus, we propose to train our model with $256{\times}256$ patches, as customary in Document Binarization, but use larger patches at inference time. This allows obtaining more training samples from the same annotated document and seamlessly exploiting more global information at inference time for better binarizing the patch. 
We demonstrate the effectiveness of our proposed approach and its superiority with respect to both standard Convolution-based and Transformer-based approaches via extensive experimental analysis on multiple benchmark datasets, whose contained documents exhibit a number of different degradations.  We release code and checkpoints at \url{https://github.com/aimagelab/FourBi}.

\section{Related Work}\label{sec:related}
\tit{Heuristics-based Approaches}
Among the strategies that exploit a handcrafted model of the image appearance, the most commonly applied is thresholding, which consists in determining a global threshold~\cite{otsu1979threshold, barron2020generalization} or a local adaptive threshold to deal with the non-uniform background typical of document images~\cite{sauvola1997adaptive}. Note that hybrid thresholding algorithms exist, such as Wolf~\cite{wolf2002text}, which use both local and global statistics. Other binarization paradigms are based on energy minimization~\cite{xiong2018historical, xiong2021enhanced} or on edge detection for modeling strokes~\cite{dang2021document, yang2023gdb}.

\tit{Learning-based Approaches} Most learning-based approaches to Document Binarizationment exploit Convolutional Neural Networks. These are employed as denoising autoencoders~\cite{zhao2018skip,he2019deepotsu} or as encoder-decoders to perform Image Processing~\cite{kang2021complex, dey2021light} or Segmentation~\cite{tensmeyer2017document, vo2018binarization}. In this work, we treat Document Binarization as a Segmentation task and devise a U-Net-like model~\cite{ronneberger2015unet} as convolutional architecture,  but with the novelty of featuring FFC layers.
Another successful research line entails treating the Document Binarizationment task as a generative one by applying Conditional GANs to obtain a binarized image starting from a degraded one in a style-transfer fashion~\cite{zhao2019document, souibgui2020gan, tamrin2021two, souibgui2021conditional}. In the context of Document Image Binarization, GANs are also employed to alleviate the training data scarcity issue connoting the task. In particular, these models are used for realistic training data augmentation~\cite{bhunia2019improving} or to achieve self-supervision via unpaired Image-to-Image Translation with Cycle-GANs~\cite{zhu2017unpaired}. 
Recent preliminary approaches employ Diffusion Models~\cite{han2023diffusion, yang2023docdiff, guo2019nonlinear} in a similar way.
To exploit long-range dependencies in combination with local information, which is difficult to achieve with standard convolutions, recent approaches~\cite{souibgui2022docentr, souibgui2022text,yang2023novel} resort to Vision Transformer (ViT)-based architectures~\cite{dosovitskiy2020image}. The encouraging performance of such methods suggests that combining global and local information is a promising strategy for Document Binarization. In this work, we explore this strategy by proposing to apply FFCs, which, contrary to standard convolutions, can model both global and local dependencies and are less computationally demanding than ViTs.

Regardless of the technique applied, the performance can be enhanced with some post-processing steps. Typically, these consist in applying morphological operators (\eg~Connected Component Analysis, Erosion, Dilation, Shrink, and Swell filtering)~\cite{lu2010document,messaoud2012region} or ensembling the output of multiple binarization algorithms~\cite{howe2013document, moghaddam2013unsupervised,akbari2019binarization}. 
In this work, we combine the output of our model run on overlapping patches of the document of interest to reduce the error on the borders, where artifacts due to discontinuity are most evident.

\tit{Frequency Information for Document Image Binarizationment} 
Previous approaches~\cite{nafchi2014phase,akbari2020binarization,lin2022three} have successfully leveraged the spectral information of documents images to address the difficulty of recognizing and recovering structures such as edges, strokes, and corners. In recent works, the Discrete Wavelet Transform~\cite{mallat1989theory}, successfully applied to Super-Resolution~\cite{zhong2018joint}, has also been used for Document Binarization to represent the document image in different frequency sub-bands either to perform Segmentation~\cite{akbari2020binarization} or to obtain a ground truth for a generator in an adversarial training setting~\cite{lin2022three}.
The main component of our architecture is the FFC, which exploits the Fast Fourier Transform (FFT). 
Moreover, unlike previous works, we do not use spectral information to represent the images but as an operator inside the architecture.

\section{Proposed Approach}\label{sec:method}
\begin{figure*}[t]
    \centering
    \includegraphics[width=\textwidth]{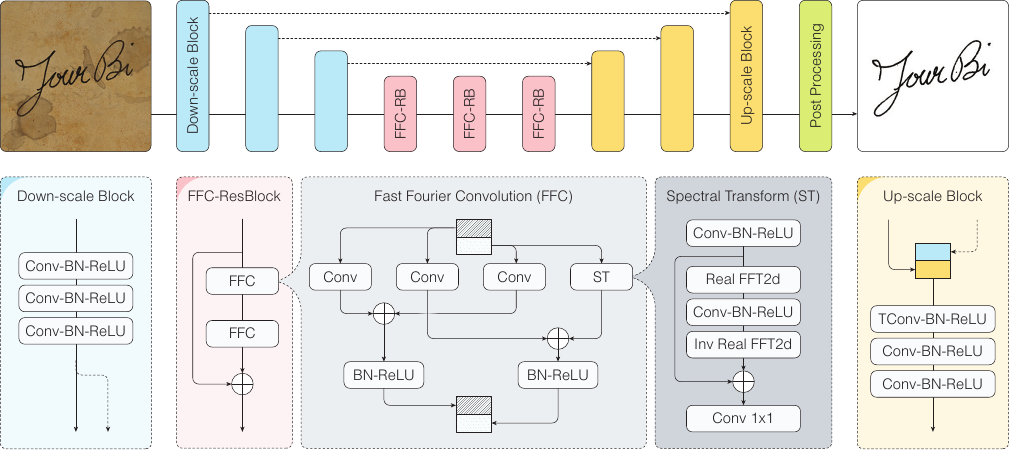}
    \caption{Overview of our Document Binarization approach exploiting FFCs (\netname). The architecture consists in a U-Net-like model in which the outputs of the down-scaling and the up-scaling blocks are combined via concatenation. The central part of the model features FFC Residual Blocks (FFC-RB). The FFC operator in these blocks combines local spatial information with global frequency information, which results in semantic-preserving precise binarization.}
    \label{fig:approach}\vspace{-1.5em}
\end{figure*}

In this work, we treat the Document Binarization task as a two-class segmentation problem, with a high imbalance towards the background class and high intra-class variability. To tackle the task, we devise a novel approach consisting of a U-Net-like model featuring FFC layers, which we call \emph{FourBi} (see~\cref{fig:approach}).

\subsection{\netname~Architecture}\label{ssec:architecture}
The proposed model takes as input square image patches, which are down-scaled via three Down-scale Blocks. Each Down-scale Block contains three sub-blocks, consisting of a Convolutional layer (Conv), a Batch Normalization (BN) layer, and a Rectified Linear Unit activation function (ReLU). The output of the last Down-scale Block is fed to a stack of FFC Residual Blocks (FFC-RB), which apply the FFC operator described below. The feature map from the last FFC-RB is then up-scaled via three Up-scale Blocks. Each of these blocks takes as input the output of the preceding Up-scale Block concatenated with the output of the corresponding Down-scale Block. The so-obtained tensor is fed to a Transpose Convolution layer (TConv), followed by a BN layer, a ReLU, and two other Conv-BN-ReLU blocks. 
The predicted tensor contains continuous values ranging from 0 to 1, representing the probability of each pixel being background or ink. From this, we obtain a binarized image by applying a threshold set equal to 0.5.

\tit{Fast Fourier Convolution for Document Binarization}
To obtain a high-quality binarized version of a document, we need the global context of the whole image in order to better model the patterns connoting it. In fact, the paper texture, the ink intensity, the color artifacts of the image scan, and the text orientation are aspects that change from document to document but are roughly similar in different parts of the same document page. This must be taken into account to distinguish the ink pixels from the background ones. 
However, working with document images also requires precision when binarizing fine parts like strokes and characters in order not to change the semantic content of the text in the document. In light of this, the global context has to be mixed with local information to classify the single pixel correctly. 
Therefore, we propose to resort to the FFC operator proposed in~\cite{chi2020fast}, which divides the network stream into a global and a local branch. The global branch exploits the Fast Fourier Transform (FFT) to enlarge the receptive field, while the local branch handles finer local patterns. 
In particular, the input tensor to the FFC layer is split into two chunks along the channel dimension, one for the local branch and one for the global branch. This way, the two branches can focus on different aspects, and different information can be stored in different parts of the tensor.
The local branch consists of two Convolutional layers, while the global branch contains a Convolutional layer and a Spectral Transform (ST) component.
The ST block exploits the two-dimensional Real FFT operator (Real FFT2d) and applies a convolution in the frequency domain. Note that the Real FFT is generally faster than the classical FFT for real-valued signals because it takes advantage of the Hermitian symmetry property. Specifically, the block works as follows. 
Given an input tensor $\mathbf{X}$ with shape $[H{\times}W{\times}C]$, it is applied the Real~FFT2d:
\begin{equation*}
\text{Real~FFT2d}: \mathbb{R}^{H{\times}W{\times}C}{\rightarrow}\mathbb{C}^{H{\times}\frac{W}{2}{\times}C},
\end{equation*}
to obtain a complex-valued tensor $\mathbf{Z}$, \ie~a tensor containing tuples that represent the real part $\mathbf{R}$ and the imaginary part $\mathbf{I}$ of the input:
\begin{equation*}
    \text{Real~FFT2d}(\mathbf{X}) = \mathbf{Z} = \mathbf{R} + i\mathbf{I}.
\end{equation*}
Then, the values in these tuples are stacked to form a real-valued tensor, \ie\\
$\mathbb{C}^{H{\times}\frac{W}{2}{\times}C}{\rightarrow}\mathbb{R}^{H{\times}\frac{W}{2}{\times}2C}$. This tensor is fed to a Convolutional block featuring a Convolutional layer with kernel size $1{\times}1$:
\begin{equation*}
\text{Conv}_{1{\times}1}\circ\text{BN}\circ\text{ReLU}: \mathbb{R}^{H{\times}\frac{W}{2}{\times}2C}{\rightarrow}\mathbb{R}^{H{\times}\frac{W}{2}{\times}2C}.
\end{equation*}
Finally, the output of this block is reshaped again into a complex-valued tensor, \ie~$\mathbb{R}^{H{\times}\frac{W}{2}{\times}2C}{\rightarrow}\mathbb{C}^{H{\times}\frac{W}{2}{\times}C}$ and fed to a two-dimensional Inverse Real FFT operator (Inv~Real~FFT2d):
\begin{equation*}
    \text{Inv~Real~FFT2d}: \mathbb{C}^{H{\times}\frac{W}{2}{\times}C}{\rightarrow}\mathbb{R}^{H{\times}W{\times}C}.
\end{equation*}
The output of the Convolutional layer in the global branch is summed with the output of one of the convolutions in the local branch, and the output of the ST is summed to the output of the other. The resulting tensors are fed to separate BN-ReLU blocks and then stacked to obtain the output of the FFC block.
Thanks to the combination of local spatial information and global frequency information, the FFC allows handling and distinguishing global pseudo-periodic patterns, such as those typical of the paper support and of the text lines, without disregarding local patterns that need to be taken into account when binarizing detailed parts of the image such as the strokes. 

\subsection{Training and Inference}
Since obtaining the ground truth information for Document Binarizationment is costly and time-consuming, publicly available datasets for the task usually contain only a few annotated pages. Therefore, as customary for this task, we enlarge the training set by training on 256${\times}$256 image patches. 
We train our model with the Charbonnier loss~\cite{charbonnier1994two}, which approximates the Mean Square Error (MSE) loss when the prediction error is smaller than the hyperparameter $\epsilon$ and the Mean Absolute Error (MAE) loss when the error is bigger~\cite{barron2019general}. Note that both the MSE and MAE loss are usually employed in Document Binarizationment~\cite{zhao2018skip,zhao2019document,bhunia2019improving,dey2021light,souibgui2022docentr,yang2023gdb}. For this reason, we adopt this loss that has been found more beneficial than these two in tasks such as image Super-Resolution~\cite{lai2018fast}. 
The loss is:
\begin{equation*}
    \mathcal{L} = \sqrt{(I - I')^2 + \epsilon^2},
\end{equation*}
where $I$ and $I'$ are the ground truth and the predicted image, respectively, and the hyperparameter $\epsilon$ is set equal to 10$^{-6}$.

At inference time, we exploit the robustness of the multi-scale representations learned by FFCs when applied to test images larger than the training ones and feed the model with 512$\times$512 patches from the test images. The increased global context helps the model distinguish between different degradation types and ink. 

\subsection{Post-Processing Strategy}\label{ssec:postprocessing}
\begin{figure}[t]
    \centering
    \includegraphics[width=.8\linewidth]{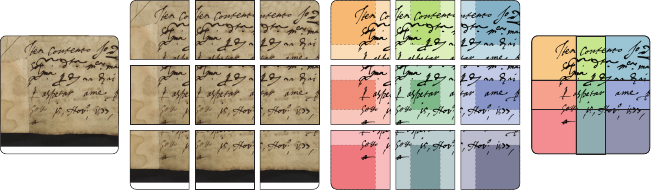}
    \caption{We combine the predictions of our model on overlapping patches at each point based on the distance between the point of interest and the center of each patch so that the artifacts at the patches borders are reduced.}
    \label{fig:overlap}\vspace{-1.5em}
\end{figure}

In the commonly adopted setting, also applied in this work, the visual quality enhancement of an entire document image is performed by working on separate local patches and then recomposing them. 
This leads to most of the errors occurring on the borders, causing discontinuities and artifacts in the reconstructed document image. 
Therefore, to enhance the performance of our approach, we introduce overlap between the patches used for prediction and keep only the innermost predictions of the model. Specifically, at inference time, we run our model over overlapping patches and recombine their prediction of each pixel by assigning to that pixel the value that it assumes in the patch to whose center it is closer. The process is shown in~\cref{fig:overlap}. 
In our experiments, we set the overlap to 256 (half the patch size), which has been determined empirically. 

\section{Experimental Analysis}\label{sec:experiments}

\tit{Implementation Details} 
Before training, we extract $384{\times}384$ image patches from the entire document image, with stride $192$ pixels. Then, during training, to further enlarge the training set, 
we apply a random rotation of \textpm 10$^{\circ}$ to these patches, and random resize and crop to obtain the $256{\times}256$ patches for training. Finally, randomly alter the image brightness, contrast, saturation, and hue with factor 0.5.
We train the model by using the AdamW optmizer~\cite{loshchilov2017decoupled} with $\beta_1{=}0.9$, $\beta_2{=}0.95$, batch size 8, and weight decay 0.05. Moreover, we set the learning rate to $1.5{\cdot}10^{-4}$ and use a cosine scheduler with 10 linear warm-up steps. 

\tit{Considered Datasets}
Various datasets have been proposed in the literature for the Document Binarization task. However, they all contain only a small number of documents because of the complexity of producing the ground-truth annotation masks. This poses a challenge when training deep learning networks, as these require large amounts of data to model the several possible degradations affecting the documents. Moreover, it is important that the set of documents used in training features a number of diverse degradations, text style, ink, and background. For this reason, in this work, we consider a number of publicly available datasets containing both handwritten and typewritten documents, from different periods and with different paper supports and degradations. 
In particular, we use the ten public benchmark datasets released over the years for the Document Image Binarization Contest (DIBCO), namely DIBCO09~\cite{gatos2009dibco09}, H-DIBCO10~\cite{pratikakis2010dibco10}, DIBCO11~\cite{pratikakis2011dibco11}, H-DIBCO12~\cite{pratikakis2012dibco12}, DIBCO13~\cite{pratikakis2013dibco13}, H-DIBCO14~\cite{ntirogiannis2014dibco14}, H-DIBCO16~\cite{pratikakis2016dibco16}, DIBCO17~\cite{pratikakis2017dibco17}, H-DIBCO18~\cite{pratikakis2018dibco18}, and DIBCO19~\cite{pratikakis2019dibco19}, each containing from 10 to 20 document images, for a total of 139. The documents feature several degradations representing real-world cases, such as variable background intensity, shadows, smear, smudge, low-contrast, bleed-through, or show-through. The documents in these datasets contain machine-printed, typed, and handwritten text in the Latin alphabet, except for DIBCO19, which also contains non-Latin characters. Note that these datasets are from disjoint sets of documents and feature different types and levels of degradation. 
Moreover, our training set also includes images from the Bickley Diary~\cite{deng2010binarizationshop}, Synchromedia Multispectral Ancient Document Images Dataset (SMADI)~\cite{hedjam2013ground}, and Nabuco~\cite{lins2011nabuco} datasets. These datasets contain documents in the Latin alphabet, both handwritten and typewritten, thus enforcing the variability of our training set.

For evaluation, we consider three modern datasets (DIBCO11, DIBCO13, DIBCO17) and five historical ones (H-DIBCO10, H-DIBCO12,  H-DIBCO14, H-DIBCO16, H-DIBCO18), and adopt the leave one-out strategy by using all other DIBCO datasets for training. 
Note that we use DIBCO19 only for testing because the patterns of the characters are too different from Latin scripts. Thus, we can evaluate the generalization capabilities of our approach to the non-Latin alphabet domain. 
In this respect, we also evaluate the performance of our model and some competitors on the non-Latin scripts Persian Heritage Image Binarization Dataset (PHIBD)~\cite{nafchi2013efficient, ayatollahi2013persian} and Irish Script on Screen Bleed-Through Database (ISOS-BTD)~\cite{rowley2012ground} datasets.

\tit{Evaluation Scores}
We evaluate our models by using the F-Measure (FM), the pseudo-FMeasure (pFM)~\cite{ntirogiannis2012performance}, the Peak Signal-to-Noise Ratio (PSNR), which measures the similarity of two images, and the Distance Reciprocal Distortion (DRD)~\cite{lu2004distance}, which measures the visual distortion of binary images. 

\subsection{Ablation Analysis}
\begin{figure*}[t]
    \centering
    \resizebox{\linewidth}{!}{
    \begin{tabular}{cc}
        \includegraphics{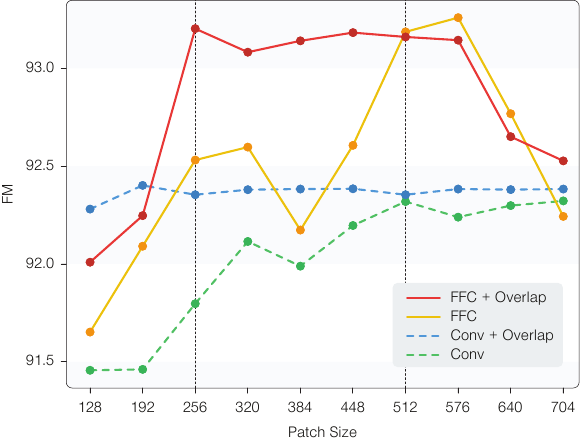} &
        \includegraphics{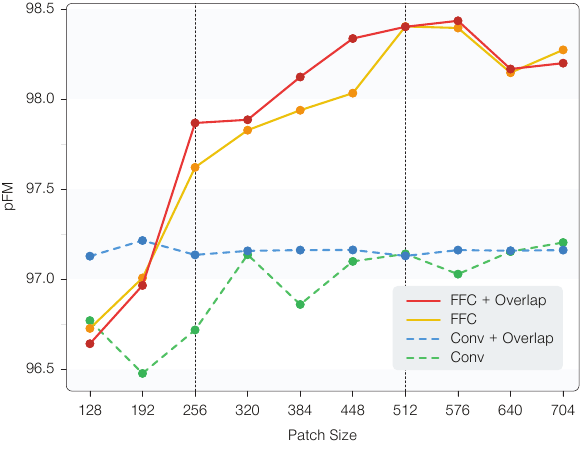} \\
        \includegraphics{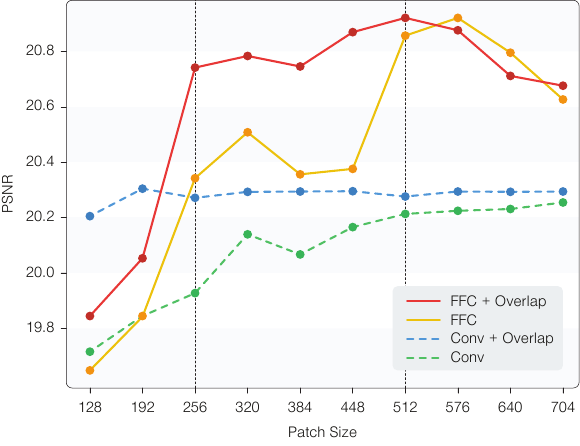} &
        \includegraphics{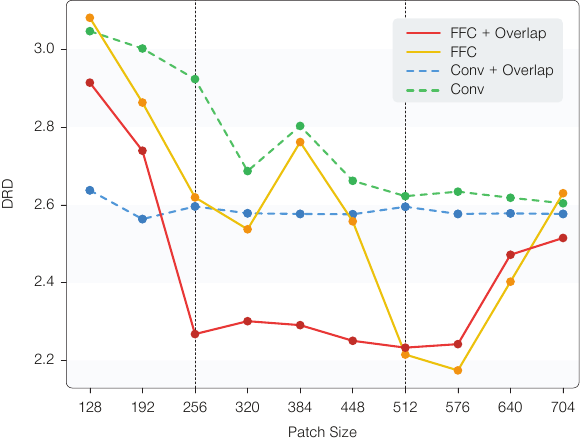} 
    \end{tabular}}
    \caption{FM (top-left), pFM (top-right), PSNR (bottom-left), and DRD (bottom-right) values obtained by our FFC-based model and a baseline variant exploiting only standard convolutions when tested on patches of different sizes from H-DIBCO18, with or without overlap.}
    \label{fig:additional_patch_size}\vspace{-1.5em}
\end{figure*}
\tit{Advantages of the FFC over the Standard Convolution}
To assess the effectiveness of FFCs over standard convolutions (Convs), we define a baseline with the same architecture as \netname~that contains solely Convs. We swap the ST module with a 1${\times}$1 convolution to maintain the global-local branch division. The 1${\times}$1 convolution, in contrast to the ST module, focuses on local aspects of the input. Therefore, the Conv-based baseline has the global and local branches logically swapped, where the global branch has the 3${\times}$3 convolutions, and the local one has the 1${\times}$1 convolution.~\cref{fig:additional_patch_size} shows the performance comparison of the two models. As we can see, the global receptive field of the FFC allows the model to outperform the Conv-based baseline.

\tit{Role of the Test Patch Size}
We ran an ablation analysis to assess the capability of our model to learn robust representations and exploit the global context provided by larger patches at inference time. In particular, we evaluate our model and the corresponding Convolutional baseline on patches whose size ranges from $128{\times}128$ to $704{\times}704$ (see~\cref{fig:additional_patch_size}). 
We observe that the scores get better consistently, reaching the best value around the patch size of 512. 

\begin{figure}[t]
\begin{minipage}[b]{0.475\linewidth}
\centering
\includegraphics[width=.76\linewidth]{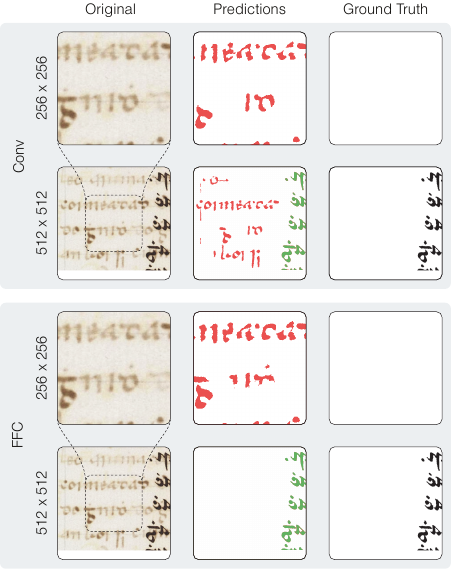}
    \caption{\strut Results of the Conv-based baseline and our FFC-based model on $256{\times}256$ and $512{\times}512$ patches. 
    Correct predictions are in green, errors in red.}
    \label{fig:patch_sizes_vs_errors}
\end{minipage}
\hfill
\begin{minipage}[b]{0.475\linewidth}
\centering
\includegraphics[width=.91\linewidth]{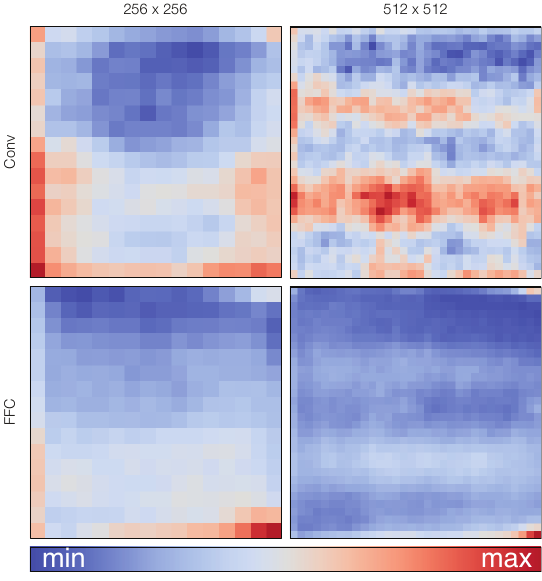}
    \caption{\strut Average error distribution of our FFC-based model and the Conv-based baseline on multiple $256{\times}256$ and $512{\times}512$ patches from different datasets. }
    \label{fig:error_on_patch}
\end{minipage}
\end{figure}

The same behavior emerges also in~\cref{fig:patch_sizes_vs_errors}, which shows how the Conv-based approach does not exploit the global information of the whole input patch and processes the images locally due to a limited receptive field. On the other hand, \netname, thanks to the FFC and its infinite receptive field, increases the performance on increasing the input patch size. From~\cref{fig:patch_sizes_vs_errors} it can be also observed that the local patch contains blurred text, which both the FFC-based and Convolution-based models incorrectly classify as foreground. In the lower part of the figure, we can see that with a larger patch, the FFC-based network leverages the surrounding context to identify that the innermost patch contains bleed-through text and correctly classifies it as background. Conversely, the Convolutional baseline does not benefit from the increased patch size and still misclassifies the blurred text as foreground.

Finally, in ~\Cref{tab:patch_sizes_docentrvsffc}, we compare the capabilities of \netname~and the Transformer-based SotA DocEnTr~\cite{souibgui2022docentr} to model long-range dependencies. When DocEnTr~\cite{souibgui2022docentr} is trained with larger patch sizes, the performance worsens, showing that the model has difficulties capturing fine-grained details of the documents. On the other hand, \netname, which already outperforms DocEnTr on the patch size 256${\times}$256, benefits from working on larger patches, even at test time only. 

\begin{table}[t]
    \footnotesize
    \centering
    \setlength{\tabcolsep}{.25em}
    \resizebox{.8\linewidth}{!}{
    \begin{tabular}{l c cl c c c c}
    \toprule
    &   \makecell{\textbf{Training}\\ \textbf{patch}} & \makecell{\textbf{Inference}\\ \textbf{patch}} &  \textbf{FM$\uparrow$} & \textbf{pFM$\uparrow$} & \textbf{PSNR$\uparrow$} & \textbf{DRD$\downarrow$} \\
    \midrule
    DocEnTr~\cite{souibgui2022docentr} & \multirow{ 2}{*}{256${\times}$256} & \multirow{ 2}{*}{256${\times}$256} & 92.53 & 95.15 & 19.11 & 2.37 \\
    \netname                           &                                  &                                  & \textbf{92.94} & \textbf{97.29} & \textbf{19.26} & \textbf{2.32} \\
    \midrule
    DocEnTr~\cite{souibgui2022docentr} & 512${\times}$512                   & \multirow{ 3}{*}{512${\times}$512} & 92.20 & 94.93 & 18.91 & 2.45 \\
    \netname                           & 256${\times}$256                   &                                  & \textbf{93.38} & \textbf{97.29} & \textbf{19.46} & \textbf{2.16} \\
    \netname                           & 512${\times}$512                   &                                  & \textit{93.27} & \textit{97.01} & \textit{19.36} & \textit{2.21} \\
    \bottomrule
    \end{tabular}}
    \caption{Performance comparison of a Transformer-based model against our FFC-based model on DIBCO17. Best results are in bold, second-best in italic.}
    \vspace{-.75cm}
    \label{tab:patch_sizes_docentrvsffc}
\end{table}

\tit{Benefits of the Post-Processing Strategy}
We perform a study of the most common errors induced by working on single patches (see~\cref{fig:error_on_patch}). In particular, for a given patch size, we evaluate the FFC-based model and the Convolutional baseline on several datasets and compute the average absolute error for each pixel location. Then, we discretize the resulting image in 16${\times}$16 bins and compute the average error distribution. The values are independently normalized for visualization purposes. From the figure, we observe that the FFC-based model is more accurate on the borders with a larger patch size. The global receptive field allows it to consider the whole information of the patch. On the other hand, the Convolutional baseline is less precise in the predictions when fed with a larger patch, and the increased error compromises the potential benefit gained from a bigger context. 
\cref{fig:additional_patch_size} also highlights that the applied overlapping procedure leads to better performance in almost all patch sizes. In particular, we notice that on the Conv-base baseline, the post-processing procedure alone saturates the performance. This behavior suggests that the main weakness of the Conv-based baseline is the predictions close to the edges. Therefore, with the overlap procedure, the number of border pixels is minimized. 

\tit{Alternative Loss Functions}\label{sec:losses}
In ~\Cref{tab:losses}, we report a comparison of the performance obtainable by training our model with other loss functions commonly adopted for Document Binarizationment. In particular, we consider the Binary Cross-Entropy (BCE), the MAE, and the MSE as alternatives to the Charbonnier loss that we adopt. We recall that the Charbonnier loss behaves as the MAE for high errors and as the MSE for low errors. 

From the table, it emerges that the error-dependant behavior of the Charbonnier loss benefits the training and leads to the best performance in terms of all metrics apart from the FM, in 
terms of which it leads to second-best results (comparable to those obtained when training with the BCE loss). Note that, for Document Binarizationement, the pFM is more relevant than the FM since it models the readability of the binarized document by penalizing more the errors made near the ground truth ink strokes. In light of this, we can conclude that the Charbonnier loss is a suitable objective function also for this task.

\begin{table}[t]
    \footnotesize
    \centering
    \resizebox{0.5\linewidth}{!}{
    \begin{tabular}{ll c c c c}
    \toprule
    && \textbf{FM$\uparrow$} & \textbf{pFM$\uparrow$} & \textbf{PSNR$\uparrow$} & \textbf{DRD$\downarrow$} \\
    \midrule
    BCE  && \textbf{93.38} & 97.18  & \textit{20.63}  & \textit{2.37} \\ 
    MSE  && 92.68  &  97.40 &  20.56 &  2.56 \\ 
    MAE  && 91.83 & \textit{98.00}  & 20.27  & 2.74 \\ 
    Charbonnier && \textit{93.19} & \textbf{98.41} &  \textbf{20.86} &  \textbf{2.21} \\
    \bottomrule
    \end{tabular}}
    \caption{Ablation analysis when training FourBi with different loss functions on the H-DIBCO18 dataset. Best results in bold, second-best in italic.}
    \label{tab:losses}
    \vspace{-1.95em}
\end{table} 

\tit{Computational Complexity}\label{sec:complexity}
Finally, we analyze the computational complexity of our approach. Note that, given a square image of size $n$ and $C_{i}$ channels, the Real FFT2d and the Inv Real FFT2d operations have time complexity $O(C_{i}{\cdot}n^2{\cdot}\text{log}n)$. The convolution has time complexity $O(C_{i}{\cdot}C_{o}{\cdot}n^2{\cdot}k^2)$, where $C_{o}$ is the output channels and $k$ is the kernel size. 
\noindent The self-attention has time complexity $O(n^2{\cdot}d)$, where $d$ is the representation dimension. 
Nevertheless, the time complexity is not a complete indicator when comparing completely different architectures.  In fact, the inductive biases of each operator, its characteristics, and how much its computation can be parallelized influence how deep and big the network needs to be in order to obtain a certain score. Therefore, to analyze the efficiency of our model, in~\Cref{tab:times}, we report runtimes obtained by processing a single image on a single NVIDIA 2080Ti GPU and the number of parameters of the considered models fed with $256{\times}256$ patches. We report this information for our approach, the baseline exploiting standard convolution, and two SotA solutions, one based on a fully-convolutional GAN (DE-GAN~\cite{souibgui2020gan}) and the other on a ViT (DocEnTr~\cite{souibgui2022docentr}).

\begin{table}[]
    \footnotesize
    \centering
    \vspace{-1.25em}
    \resizebox{.4\linewidth}{!}{
    \begin{tabular}{l r r}
    \toprule
    & \textbf{params} & \textbf{runtime} \\
    \midrule
    DE-GAN~\cite{souibgui2021conditional} & 31.0M & 11.2 ms \\
    DocEnTr~\cite{souibgui2022docentr}    & 68.8M &  3.8 ms \\
    Conv. Baseline                        & 15.8M &  4.7 ms \\
    FourBi                                & 17.6M &  9.1 ms \\  
    \bottomrule
    \end{tabular}}
    \caption{Computational complexity and performance of SotA approaches and the Conv-based baseline on the H-DIBCO18 dataset, all fed with $256{\times}256$ patches.}
    \label{tab:times}
    \vspace{-3.85em}
\end{table}

\subsection{Comparison with the SotA}
\tit{Quantitative comparison on Latin-characters documents}
We evaluate our method on eight DIBCO datasets. The results are reported in~\Cref{tab:res_dibco10,tab:res_dibco11,tab:res_dibco12,tab:res_dibco13,tab:res_dibco14,tab:res_dibco16,tab:res_dibco17,tab:res_dibco18}. For space reasons, we sorted the competitors by PSNR, and we kept for each table the first top ten methods. Moreover, to better highlight the consistency of our method, we include for each table the \textit{rank} of~\netname~compared to the competitors.
It can be observed that our method achieves top performance on most metrics. In the tables, we isolate Transformer-based methods, which share with our approach the capability to model both long-range and short-range dependencies. Further, our model is adaptable to different image resolutions and has no strict constraints on the input size. 

\begin{table}[t]
\begin{minipage}[b]{.49\textwidth }
\footnotesize
    \centering
    \resizebox{\linewidth}{!}{
    \begin{tabular}{ll c c c c}
			\toprule
                &   & \textbf{FM$\uparrow$} & \textbf{pFM$\uparrow$} & \textbf{PSNR$\uparrow$} & \textbf{DRD$\downarrow$} \\
			\midrule
			Winning Entry~\cite{pratikakis2010dibco10}       &   & 91.50                 & 93.58                  & 19.78                   & -                        \\
			\midrule
			Guo \etal~\cite{guo2019nonlinear}                &   & 86.75                 & 89.74                  & 17.93                   & 3.62                     \\
			Zhang \etal~\cite{zhang2020selective}            &   & 85.70                 & 90.61                  & 17.96                   & 3.60                     \\
			Feng~\cite{feng2022effective}                    &   & 88.16                 & 90.96                  & 18.25                   & 3.41                     \\
			Du \& He~\cite{du2021nonlinear}                  &   & 88.47                 & 90.87                  & 18.41                   & 3.22                     \\
			Guo \etal~\cite{guo2020fourth}                   &   & 88.56                 & 93.50                  & 18.56                   & 3.06                     \\
			Jia~\etal~\cite{jia2018degraded}                 &   & 91.75                 & 95.36                  & 19.84                   & 2.17                     \\
			Suh~\etal~\cite{suh2020two}                      &   & 93.92                 & 96.53                  & 21.18                   & 1.50                     \\
			cGANs~\cite{zhao2019document}                    &   & 93.92                 & 96.53                  & 21.18                   & 1.50                     \\
			GDB~\cite{yang2023gdb}                           &   & 95.19                 & 96.62                  & 21.98                   & 1.32                     \\
			\midrule
			D\textsuperscript{2}BFormer~\cite{yang2023novel} &   & 96.23                 & 97.49                  & 23.10                   & 0.95                     \\
			\midrule
			\netname                                         &   & \textbf{96.43}        & \textbf{98.60}         & \textbf{23.37}          & \textbf{0.92}            \\
			Rank && 1\textsuperscript{st} & 1\textsuperscript{st} & 1\textsuperscript{st} & 1\textsuperscript{st} \\
			\bottomrule
		\end{tabular}}
    \caption{\strut Quantitative comparison on the H-DIBCO10 dataset. 
    }
    \label{tab:res_dibco10} 
\hrule height 0pt
\end{minipage}
\hfill
\begin{minipage}[b]{.49\textwidth}
\footnotesize
    \centering
    \resizebox{\linewidth}{!}{
        \begin{tabular}{ll c c c c}
			\toprule
			                                                 &   & \textbf{FM$\uparrow$} & \textbf{pFM$\uparrow$} & \textbf{PSNR$\uparrow$} & \textbf{DRD$\downarrow$} \\
			\midrule
			Winning Entry~\cite{pratikakis2011dibco11}       &   & 88.74                 & -                      & 17.97                   & 5.36                     \\
			\midrule
			Jia~\etal~\cite{jia2018degraded}                 &   & 91.65                 & 95.56                  & 18.88                   & 2.66                     \\
			HDSN~\cite{vo2018binarization}                   &   & 92.58                 & 94.67                  & 19.16                   & 2.38                     \\
			Suh~\etal~\cite{suh2020two}                      &   & 93.44                 & 96.18                  & 19.97                   & 1.93                     \\
			GDB~\cite{yang2023gdb}                           &   & 93.44                 & 95.82                  & 20.01                   & 2.25                     \\
			cGANs~\cite{zhao2019document}                    &   & 93.81                 & 95.70                  & 20.26                   & 1.81                     \\
			Lin~\etal~\cite{lin2022three}                    &   & 94.08                 & 97.08                  & 20.51                   & 1.75                     \\
			Dang~\etal~\cite{dang2021document}               &   & 95.61                 & 97.34                  & 22.09                   & 1.48                     \\
			\midrule
			DocEnTr~\cite{souibgui2022docentr}               &   & 94.37                 & 96.15                  & 20.81                   & 1.63                     \\
			D\textsuperscript{2}BFormer~\cite{yang2023novel} &   & 94.82                 & 96.62                  & 21.27                   & 1.56                     \\
			Text-DIAE~\cite{souibgui2022text}                &   & 95.01                 & 96.86                  & 21.29                   & 1.48                     \\
			\midrule
			\netname                                         &   & \textbf{96.03}        & \textbf{98.70}         & \textbf{22.26}          & \textbf{1.19}            \\
			Rank && 1\textsuperscript{st} & 1\textsuperscript{st} & 1\textsuperscript{st} & 1\textsuperscript{st} \\
			\bottomrule
		\end{tabular}}
    \caption{\strut Quantitative comparison on the DIBCO11 dataset. }
    \label{tab:res_dibco11}
\hrule height 0pt
\end{minipage}\vspace{-1.5em}
\end{table}
\begin{table}[t]
\begin{minipage}[b]{.49\textwidth }
\footnotesize
    \centering
    \resizebox{\linewidth}{!}{
    \begin{tabular}{ll c c c c}
			\toprule
			                                                 &   & \textbf{FM$\uparrow$} & \textbf{pFM$\uparrow$} & \textbf{PSNR$\uparrow$} & \textbf{DRD$\downarrow$} \\
			\midrule
			Winning Entry~\cite{pratikakis2012dibco12}       &   & 92.85                 & -                      & 21.80                   & 2.66                     \\
			\midrule
			Du \& He~\cite{du2021nonlinear}                  &   & 89.40                 & 91.16                  & 18.80                   & 3.66                     \\
			Jia~\etal~\cite{jia2018degraded}                 &   & 92.96                 & 95.76                  & 20.43                   & 2.30                     \\
			Suh~\etal~\cite{suh2020two}                      &   & 94.50                 & 97.36                  & 21.78                   & 1.73                     \\
			cGANs~\cite{zhao2019document}                    &   & 94.96                 & 96.15                  & 21.91                   & 1.55                     \\
			Jemni~\etal~\cite{jemni2022enhance}              &   & 95.18                 & 94.63                  & 22.00                   & 1.62                     \\
			GDB~\cite{yang2023gdb}                           &   & 95.80                 & 97.03                  & 22.62                   & 1.32                     \\
			\scriptsize{Rezabezhad~\etal~\cite{rezanezhad2023hybrid}}     && 96.25  & 97.58 & 23.27   & 1.11           \\
			\midrule
			DocEnTr~\cite{souibgui2022docentr}               &   & 95.31                 & 96.29                  & 22.29                   & 1.60                     \\
			D\textsuperscript{2}BFormer~\cite{yang2023novel} &   & 96.28                 & 96.90                  & 23.35                   & 1.26                     \\
			Text-DIAE~\cite{souibgui2022text}                &   & 96.52                 & 97.04                  & 23.66                   & 1.10                     \\
			\midrule
			\netname                                         &   & \textbf{97.07}        & \textbf{98.78}         & \textbf{24.29}          & \textbf{0.94}            \\
			  Rank && 1\textsuperscript{st} & 1\textsuperscript{st} & 1\textsuperscript{st} & 1\textsuperscript{st} \\
			\bottomrule
		\end{tabular}}
    \caption{\strut Quantitative comparison on the H-DIBCO12 dataset.}
    \label{tab:res_dibco12} 
\hrule height 0pt
\end{minipage}
\hfill
\begin{minipage}[b]{.49\textwidth}
\footnotesize
    \centering
    \resizebox{\linewidth}{!}{
        \begin{tabular}{ll c c c c}
			\toprule
			                                                 &   & \textbf{FM$\uparrow$} & \textbf{pFM$\uparrow$} & \textbf{PSNR$\uparrow$} & \textbf{DRD$\downarrow$} \\
			\midrule
			Winning Entry~\cite{pratikakis2013dibco13}       &   & 92.12                 & 94.19                  & 20.68                   & 3.10                     \\
			\midrule
			Du \& He~\cite{du2021nonlinear}                  &   & 89.94                 & 93.97                  & 19.26                   & 3.24                     \\
			Jia~\etal~\cite{jia2018degraded}                 &   & 93.28                 & 96.58                  & 20.76                   & 2.01                     \\
			HDSN~\cite{vo2018binarization}                   &   & 93.43                 & 95.34                  & 20.82                   & 2.26                     \\
			cGANS~\cite{zhao2019document}                    &   & 93.86                 & 96.47                  & 21.53                   & 2.32                     \\
			Suh~\etal~\cite{suh2020two}                      &   & 94.75                 & 97.20                  & 21.79                   & 1.66                     \\
			Lin~\etal~\cite{lin2022three}                    &   & 95.24                 & 97.51                  & 22.27                   & 1.59                     \\
			GDB~\cite{yang2023gdb}                           &   & 95.19                 & 96.37                  & 22.58                   & 1.78                     \\
			Dang~\etal~\cite{dang2021document}               &   & 95.96                 & 98.13                  & 23.14                   & 1.43                     \\
			DE-GAN~\cite{souibgui2020gan}                    &   & \textbf{99.50}        & \textbf{99.70}         & \textbf{24.90}          & \textbf{1.10}            \\
			\midrule
			D\textsuperscript{2}BFormer~\cite{yang2023novel} &   & 95.96                 & 96.96                  & 22.80                   & 1.30                     \\
			\midrule
			\netname                                         &   & 96.71                 & 98.27                  & 24.17                   & 1.13                     \\
                Rank && 2\textsuperscript{nd} & 2\textsuperscript{nd} & 2\textsuperscript{nd} & 2\textsuperscript{nd} \\
			\bottomrule
		\end{tabular}}
    \caption{\strut Quantitative comparison on the DIBCO13 dataset. }
    \label{tab:res_dibco13}
\hrule height 0pt
\end{minipage}\vspace{-1.5em}
\end{table}
\begin{table}[t]
\begin{minipage}[b]{.49\textwidth }%
\footnotesize
    \centering
    \resizebox{\linewidth}{!}{
        \begin{tabular}{ll c c c c}
			\toprule
			                                                 &   & \textbf{FM$\uparrow$} & \textbf{pFM$\uparrow$} & \textbf{PSNR$\uparrow$} & \textbf{DRD$\downarrow$} \\
			\midrule
			Winning Entry~\cite{ntirogiannis2014dibco14}     &   & 96.88                 & 97.65                  & 22.66                   & 0.90                     \\
			\midrule
			Feng~\cite{feng2022effective}                    &   & 92.99                 & 95.54                  & 19.42                   & 2.26                     \\
			Du \& He~\cite{du2021nonlinear}                  &   & 93.48                 & 95.49                  & 19.80                   & 2.02                     \\
			Jia~\etal~\cite{jia2018degraded}                 &   & 94.89                 & 97.68                  & 20.53                   & 1.5                      \\
			Suh~\etal~\cite{suh2020two}                      &   & 96.19                 & 97.13                  & 21.77                   & 1.14                     \\
			cGANs~\cite{zhao2019document}                    &   & 96.41                 & 97.55                  & 22.12                   & 1.07                     \\
			Lin~\etal~\cite{lin2022three}                    &   & 86.65                 & 98.19                  & 22.27                   & 0.96                     \\
			GDB~\cite{yang2023gdb}                           &   & 97.66                 & 97.26                  & 23.13                   & 1.21                     \\
			HDSN~\cite{vo2018binarization}                   &   & 96.66                 & 97.59                  & 23.23                   & 0.79                     \\
			\midrule
			DocEnTr~\cite{souibgui2022docentr}               &   & 97.16                 & 98.28                  & 22.99                   & \textbf{0.16}            \\
			D\textsuperscript{2}BFormer~\cite{yang2023novel} &   & 97.70                 & 98.44                  & 23.91                   & 0.64                     \\
			\midrule
			\netname                                         &   & \textbf{98.19}        & \textbf{99.16}         & \textbf{25.18}          & 0.55                     \\
			Rank && 1\textsuperscript{st} & 1\textsuperscript{st} & 1\textsuperscript{st} & 2\textsuperscript{nd} \\
			\bottomrule
		\end{tabular}
    }
    \caption{\strut Quantitative comparison on the H-DIBCO14 dataset. }
    \label{tab:res_dibco14}
\hrule height 0pt
\end{minipage}
\hfill
\begin{minipage}[b]{.49\textwidth}
\footnotesize
    \centering
    \resizebox{\linewidth}{!}{
        \begin{tabular}{ll c c c c}
			\toprule
			                                                 &   & \textbf{FM$\uparrow$} & \textbf{pFM$\uparrow$} & \textbf{PSNR$\uparrow$} & \textbf{DRD$\downarrow$} \\
			\midrule
			Winning Entry~\cite{pratikakis2016dibco16}       &   & 88.72                 & 91.84                  & 18.45                   & 3.86                     \\
			\midrule
			Feng~\cite{feng2022effective}                    &   & 88.88                 & 90.30                  & 18.45                   & 4.99                     \\
			Zhang \etal~\cite{zhang2020selective}            &   & 88.04                 & 90.89                  & 18.58                   & 4.00                     \\
			HDSN~\cite{vo2018binarization}                   &   & 90.01                 & 93.44                  & 18.74                   & 3.91                     \\
			cGANs~\cite{zhao2019document}                    &   & 89.77                 & 94.85                  & 18.80                   & 3.85                     \\
			Jia~\etal~\cite{jia2018degraded}                 &   & 90.01                 & 93.72                  & 19.00                   & 4.03                     \\
			GDB~\cite{yang2023gdb}                           &   & 90.41                 & 94.70                  & 19.00                   & 3.34                     \\
			Suh~\etal~\cite{suh2020two}                      &   & 91.11                 & 95.22                  & 19.34                   & 3.25                     \\
			Du \& He~\cite{du2021nonlinear}                  &   & 91.51                 & 91.81                  & 19.62                   & 3.16                     \\
			Jemni~\etal~\cite{jemni2022enhance}              &   & \textbf{94.95}        & 94.55                  & \textbf{21.85}          & \textbf{1.56}            \\
			\midrule
			D\textsuperscript{2}BFormer~\cite{yang2023novel} &   & 90.96                 & 93.30                  & 19.24                   & 3.22                     \\
			\midrule
			\netname                                         &   & 91.30                 & \textbf{96.62}         & 19.74                   & 3.02                     \\
                Rank && 2\textsuperscript{nd} & 1\textsuperscript{st} & 2\textsuperscript{nd} & 2\textsuperscript{nd} \\
			\bottomrule
		\end{tabular}
    }
    \caption{\strut Quantitative comparison on the H-DIBCO16 dataset.}
    \label{tab:res_dibco16}
\hrule height 0pt
\end{minipage}\vspace{-1.5em}
\end{table}
\begin{table}[t]
\begin{minipage}[b]{.49\textwidth }%
\footnotesize
    \centering
    \resizebox{\linewidth}{!}{
        \begin{tabular}{ll c c c c}
			\toprule
			                                                 &   & \textbf{FM$\uparrow$} & \textbf{pFM$\uparrow$} & \textbf{PSNR$\uparrow$} & \textbf{DRD$\downarrow$} \\
			\midrule
			Winning Entry~\cite{pratikakis2017dibco17}       &   & 91.04                 & 92.86                  & 18.28                   & 3.40                     \\
			\midrule
			cGANs~\cite{zhao2019document}                    &   & 90.73                 & 92.58                  & 17.83                   & 3.58                     \\
			Suh~\etal~\cite{suh2020two}                      &   & 90.95                 & 94.65                  & 18.40                   & 2.93                     \\
			Lin~\etal~\cite{lin2022three}                    &   & 90.95                 & 93.79                  & 18.57                   & 2.94                     \\
			Dang~\etal~\cite{dang2021document}               &   & 92.08                 & 94.99                  & 18.72                   & 2.84                     \\
			DE-GAN~\cite{souibgui2020gan}                    &   & \textbf{97.71}        & \textbf{98.23}         & 18.74                   & 3.01                     \\
			Rezabezhad \etal~\cite{rezanezhad2023hybrid}     &   & 93.01                 & 95.42                  & 19.24                   & 2.29                     \\
			GDB~\cite{yang2023gdb}                           &   & 94.32                 & 96.58                  & \textbf{20.04}          & \textbf{1.79}            \\
			\midrule
			DocEnTr~\cite{souibgui2022docentr}               &   & 92.53                 & 95.15                  & 19.11                   & 2.37                     \\
			D\textsuperscript{2}BFormer~\cite{yang2023novel} &   & 93.52                 & 95.09                  & 19.35                   & 2.12                     \\
			Text-DIAE~\cite{souibgui2022text}                &   & 93.84                 & 95.71                  & 19.64                   & 1.93                     \\
			\midrule
			\netname                                         &   & 93.81                 & 96.57                  & 19.66                   & 2.03                     \\
                Rank && 4\textsuperscript{th} & 3\textsuperscript{rd} & 2\textsuperscript{nd} & 3\textsuperscript{rd} \\
			\bottomrule
		\end{tabular}
    }
    \caption{\strut Quantitative comparison on the DIBCO17 dataset. }
    \label{tab:res_dibco17}
\hrule height 0pt
\end{minipage}
\hfill
\begin{minipage}[b]{.49\textwidth}
\footnotesize
    \centering
    \resizebox{\linewidth}{!}{
		\begin{tabular}{ll c c c c}
			\toprule
			                                                 &   & \textbf{FM$\uparrow$} & \textbf{pFM$\uparrow$} & \textbf{PSNR$\uparrow$} & \textbf{DRD$\downarrow$} \\
			\midrule
			Winning Entry~\cite{pratikakis2018dibco18}       &   & 88.34                 & 90.24                  & 19.11                   & 4.92                     \\
			\midrule
			DE-GAN~\cite{souibgui2020gan}                    &   & 77.59                 & 85.74                  & 16.16                   & 7.93                     \\
			Dang~\etal~\cite{dang2021document}               &   & 91.26                 & 93.97                  & 19.81                   & 3.42                     \\
			GDB~\cite{yang2023gdb}                           &   & 91.09                 & 94.57                  & 19.92                   & 3.07                     \\
			Lin~\etal~\cite{lin2022three}                    &   & 91.66                 & 95.53                  & 20.02                   & 2.81                     \\
			Suh~\etal~\cite{suh2020two}                      &   & 91.86                 & 96.25                  & 20.03                   & 2.60                     \\
			Jemni~\etal~\cite{jemni2022enhance}              &   & 92.41                 & 94.35                  & 20.18                   & 2.60                     \\
			Rezabezhad \etal~\cite{rezanezhad2023hybrid}     &   & 92.47                 & 95.99                  & 20.29                   & 2.50                     \\
			\midrule
			D\textsuperscript{2}BFormer~\cite{yang2023novel} &   & 88.84                 & 93.42                  & 18.91                   & 3.99                     \\
			DocEnTr~\cite{souibgui2022docentr}               &   & 90.59                 & 93.97                  & 19.46                   & 3.35                     \\
			Text-DIAE~\cite{souibgui2022text}                &   & 91.32                 & 94.44                  & 19.95                   & 3.21                     \\
			\midrule
			\netname                                         &   & \textbf{93.16}        & \textbf{98.40}         & \textbf{20.92}          & \textbf{2.23}            \\
			Rank && 1\textsuperscript{st} & 1\textsuperscript{st} & 1\textsuperscript{st} & 1\textsuperscript{st} \\
			\bottomrule
		\end{tabular}}
	\caption{Quantitative comparison on the H-DIBCO18 dataset.}
	\label{tab:res_dibco18}
\hrule height 0pt
\end{minipage}\vspace{-1.5em}
\end{table}

\tit{Quantitative comparison on non-Latin-characters documents}\label{sec:quantitatives} In ~\Cref{tab:nonlatin}, we report a quantitative performance analysis on the non-Latin DIBCO19, PHIBD, and ISOS-BTD datasets, which were not included in any training set. Specifically, we compare our approach, both alone and combined with the overlap-based post-processing strategy, with two SotA approaches: the fully-convolutional, generative DE-GAN~\cite{souibgui2020gan} and the ViT-based DocEnTr~\cite{souibgui2022docentr}. 
It can be observed the same trend as for the considered Latin-characters training datasets. Specifically, FourBi achieves competitive or better performance compared to DE-GAN and DocEnTr. 
The difference in the charset prevents the networks from exploiting a priori knowledge possibly learned on the other datasets, which might allow them to recognize \eg~whether the text is flipped (bleed-through) from the context. Therefore, the networks can only exploit the image appearance to classify each pixel correctly. Nonetheless, thanks to the FFC, FourBi can exploit global information, which helps it better model the overall visual appearance.

\begin{table}[t]
    \footnotesize
    \centering
    \setlength{\tabcolsep}{.65em}
    \resizebox{.7\linewidth}{!}{
    \begin{tabular}{c ll c c c c}
    \toprule
    &&& \textbf{FM$\uparrow$} & \textbf{pFM$\uparrow$} & \textbf{PSNR$\uparrow$} & \textbf{DRD$\downarrow$} \\
    \midrule
    \parbox[t]{2mm}{\multirow{4}{*}{\rotatebox[origin=c]{90}{\textbf{PHIBD}}}}
    &DE-GAN~\cite{souibgui2020gan} && 69.83 & 71.58 & 12.33 & 28.30 \\
    &DocEnTr~\cite{souibgui2022docentr} && 59.28 & 83.81 & 13.91 & 15.89 \\
    \cmidrule{2-7}
    &\netname && \textit{80.07} & \textit{84.05} & \textit{17.31} & \textit{14.07} \\
    &\netname~+ Overlap &&\textbf{80.36}&\textbf{90.66}&\textbf{17.41}&\textbf{13.22}\\
    \midrule
    \parbox[t]{2mm}{\multirow{4}{*}{\rotatebox[origin=c]{90}{\textbf{ISOS-BTD}}}}
    &DE-GAN~\cite{souibgui2020gan} &&\textbf{85.19}&91.55&\textit{12.59}&\textbf{13.25}\\
    &DocEnTr~\cite{souibgui2022docentr} &&\textit{79.34}&\textit{97.48}&12.44&14.12\\
    \cmidrule{2-7}
    &\netname  && 78.23 & \textbf{98.75} & 12.46 & 14.12 \\
    &\netname~+ Overlap && 78.88 & \textbf{98.75} & \textbf{12.62} & \textit{13.64} \\
    \midrule    
    \parbox[t]{2mm}{\multirow{4}{*}{\rotatebox[origin=c]{90}{\textbf{DIBCO19}}}}
    &DE-GAN~\cite{souibgui2020gan}                    && 57.96 & 57.30 & 14.43 & 12.21  \\ 
    &DocEnTr~\cite{souibgui2022docentr}              && \textbf{67.20} & 77.13 & 14.12 & 10.20  \\ 
    \cmidrule{2-7}
    &\netname                                 && 63.15 & \textit{84.99} & \textit{14.94} & \textit{9.89} \\ 
    &\netname~+ Overlap              && \textit{63.56} & \textbf{85.82} & \textbf{15.02} & \textbf{9.66} \\ 
    \bottomrule
    \end{tabular}}
    \caption{Quantitative comparison on the considered non-Latin character datasets. Best results in bold, second-best in italic for each of them.}\vspace{-1.5em}
    \label{tab:nonlatin}
\end{table}

\begin{figure}[t]
    \centering
    \includegraphics[width=\linewidth]{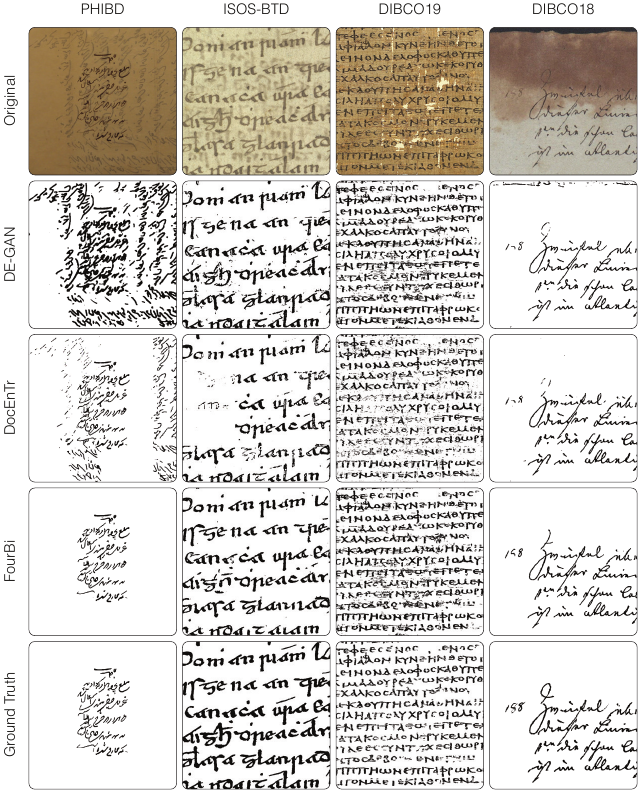}
    \caption{Qualitative comparison between~\netname~and other SotA approaches on datasets containing ancient documents. Starting from the left column, we test~\netname~on the non-Latin datasets PHIBD~\cite{nafchi2013efficient, ayatollahi2013persian}, ISOS-BTD~\cite{rowley2012ground}, and DIBCO19~\cite{pratikakis2019dibco19}. In the rightmost column, we test~\netname~on the Latin dataset DIBCO18~\cite{pratikakis2018dibco18}. All models were trained on Latin-script datasets.}
    \label{fig:qualitatives}
\end{figure}

\tit{Qualitative Comparison on with the SoTa}
We qualitatively compare our approach, DE-GAN~\cite{souibgui2020gan}, and DocEnTr~\cite{souibgui2022docentr}, on Latin and non-Latin documents. 
We show the results with one of the challenging images of H-DIBCO18~\cite{pratikakis2018dibco18} in~\cref{fig:qualitatives}-Right. As we can see, both competitors produce images with speckles on the background, while our model, thanks to its ability to manage the global information, produces less noisy predictions and recognizes the page limit on top. Nonetheless, this sample is heavily stained on the top, and all three methods fail at accurately binarizing part of the head loop on the capital letter ``G'' and the page number.
Finally, we qualitatively compare our approach on images from challenging ancient datasets.
Different from those used in training, these documents contain non-Latin text. Thus, our model would not be able to exploit any potentially learned characters appearance information when binarizing those images. In~\cref{fig:qualitatives}-Left, it can be observed that FourBi accurately removes severe degradations such as faded ink and holes. Moreover, it shows superior capabilities in handling bleed-through ink, exploiting the global context.

\section{Conclusion}\label{sec:conclusion}
In this work, we have proposed a fully-convolutional Document Binarization approach that can handle both global and local information to predict more precisely whether a pixel is background or ink. We achieve this capability by exploiting the FFC operator in a U-Net-like architecture. 
The FFC layers allow handling both local and global information, thus leading to results that are comparable or superior to those achieved with Transformer-based models.
Moreover, we have observed that the common strategy to divide the document image into patches and treat them separately leads to artifacts at the borders of the patches. In light of this, we have applied a simple yet effective post-processing strategy consisting in combining the predictions of our model on overlapping patches. 
The results of deep experimental analysis on multiple benchmark datasets demonstrate the suitability of our approach, together with the possibility of applying the FFC operator to patches that are much wider than those used in training without performance drop, further improving the performance by reducing the artifacts at the borders.

\section*{Acknowledgement}
This work was supported by the ``AI for Digital Humanities'' project (Pratica Sime n.2018.0390), funded by ``Fondazione di Modena'' and the PNRR project Italian Strengthening of ESFRI RI Resilience (ITSERR) funded by the European Union – NextGenerationEU (CUP: B53C22001770006).

\clearpage
\bibliographystyle{splncs04}
\bibliography{main}

\end{document}